\documentclass{article}

\usepackage{arxiv}
\usepackage{float}  

\usepackage[utf8]{inputenc} 
\usepackage[T1]{fontenc}    
\usepackage{hyperref}       
\usepackage{url}            
\usepackage{booktabs}       
\usepackage{amsfonts}       
\usepackage{nicefrac}       
\usepackage{microtype}      
\usepackage{lipsum}
\usepackage{graphicx}
\usepackage{pgfplots}
\usepackage{listings} 
\usepackage{color}
\usepackage{tcolorbox}
\usepackage{amsmath}
\pgfplotsset{compat=1.17} 
\usepackage{pgfplotstable} 
\graphicspath{ {./images/} }
\usepackage{tikz}
\usetikzlibrary{shapes,arrows,fit,positioning,shapes.geometric}

\title{REL: Working out is all you need}

\author{
    \begin{tabular}{c@{\hspace{2cm}}c@{\hspace{2cm}}c}
        Toby Simonds$^{1,2}$ & Jey Han Lau$^{2}$ & Chaithanya Bandi$^{1}$ \\
        \texttt{toby@withmartian.com} & \texttt{laujh@unimelb.edu.au} & \texttt{chaithanya@withmartian.com} \\[2ex]
    \end{tabular}
    \\[2ex]
    $^{1}$Martian AI \\
    $^{2}$The University of Melbourne
}

\begin{document}
\maketitle
\begin{abstract}
Recent developments, particularly OpenAI’s O1 model, have demonstrated the remarkable potential of Large Language Models (LLMs) for complex reasoning tasks. Through analysis of O1’s outputs and provided sample Chain-of-Thought (CoT) demonstrations, we observe that it approaches problem- solving in a distinctly human-like manner—systematically brainstorming ideas, testing hypotheses, verifying results, and planning comprehensive solutions. These sophisticated reasoning capabilities remain notably absent in other state-of-the-art language models. In this paper, we hypothesize that this performance gap stems from the limited availability of high-quality reasoning process data in current training sets. We demonstrate that by constructing a specialized dataset focused on explicit problem-solving workflows (``worked solutions''), we can elicit substantially improved planning capabilities from existing models. Additionally, we  propose  Reasoning Enhancement Loop (REL), a method for generating synthetic worked solutions. All code and data are released at: \url{https://github.com/tamassimonds/REL}
\end{abstract}


\section{Introduction}
OpenAI's O1 \cite{o1systemcard} represents a step-function improvement in language model capabilities, particularly in its approach to problem-solving. Analysis of O1's Chain-of-Thought (CoT) samples reveals a fundamental shift in reasoning strategy: while traditional LLMs typically produce direct solutions akin to textbook answers, O1 exhibits a more exploratory problem-solving approach that mirrors advanced human cognition—systematically investigating problem spaces, considering multiple approaches, and iteratively refining solutions.

We propose that the direct, solution-focused reasoning of traditional LLMs stems from their training data composition. Current datasets predominantly consist of problem-solution pairs rather than detailed problem-solving processes. While methods like rejection sampling and quality filtering have successfully generated additional high-quality synthetic training data, they fundamentally fail to capture the sophisticated reasoning processes exhibited by O1 \cite{valmeekam_llms_2024}. Our empirical testing reveals that current synthetic data generation approaches, despite producing correct solutions, cannot replicate O1's characteristic exploratory problem-solving behavior—its ability to brainstorm approaches, test hypotheses, and iteratively refine solutions \cite{kambhampati_llms_2024}. This limitation persists even when applying state-of-the-art filtering techniques to select the highest quality samples, suggesting that the ability to generate intermediate reasoning steps cannot be easily bootstrapped from existing models trained primarily on final solutions.

In this paper, we demonstrate that by constructing a high-quality dataset of fully ``worked solutions''—including brainstorming, hypothesis testing, and solution refinement—we can elicit fundamental reasoning capabilities in language models. This initial dataset, though resource-intensive to create, proves crucial for inducing basic planning behaviors that can then be amplified through our Reasoning Enhancement Loop (REL), a critic-generator pipeline that autonomously produces additional high-quality worked solutions. Our empirical investigation reveals that attempting to directly apply self-improvement techniques to models without this foundational working out capacity fails to produce sophisticated planning behavior. However, by first finetuning a base LLM on expert demonstrations that explicitly showcase problem exploration and solution planning, we establish the necessary scaffolding for our REL pipeline to then iteratively enhance these capabilities through autonomous generation and refinement of worked solutions.

Traditionally, collecting such comprehensive problem-solving data has been prohibitively expensive, requiring both significant time investment and advanced domain expertise (typically graduate-level mathematicians). We address this challenge by introducing a novel AI-augmented data collection methodology that combines human expertise with automated assistance, improving collection efficiency by an order of magnitude while maintaining data quality.

\paragraph{Key Contributions} Our main contributions are:
\begin{itemize}
    \item A novel dataset creation methodology that combines human expertise with AI assistance to efficiently generate high-quality problem-solving worked solutions at scale.
    
    \item ReasonSet, a dataset of worked solutions that captures the problem-solving process, including brainstorming, hypothesis testing, and solution refinement stages.
    
    \item REL: A critic-generator pipeline that autmomatically generates additional high-quality worked solutions through iterative refinment and validation.
    
    \item Empirical evidence demonstrating that models trained on worked solutions exhibit significantly improved planning and problem-solving capabilities compared to traditional approaches, e.g.\ we found a 18.9\% improvement on AIME 2024.
    \item The release of O1-Llama 3.2 3B, a proof of concept of how we can elicit such reasoning abilities in LLMs.
\end{itemize}

\section{Background and Related Work}

\subsection{Large Language Models and Reasoning}
Recent advances in Large Language Models (LLMs) have shown remarkable capabilities in complex reasoning tasks \cite{gpt4osystemcard}. Traditional approaches to improving LLM reasoning have focused on techniques like Chain-of-Thought prompting \cite{wei_chain--thought_2023} and Self-Taught Reasoning \cite{zelikman_star_2022}, which guide models to break down problems into intermediate steps. While these methods have shown success in mathematical and logical reasoning tasks, they typically produce linear solution paths rather than the exploratory problem-solving demonstrated by human experts. 

The release of O1 \cite{o1systemcard} represents a significant departure from this paradigm. Unlike previous models that generate direct solutions, O1 demonstrates a more sophisticated approach to problem-solving, actively exploring solution spaces and testing multiple hypotheses before arriving at conclusions. This capability suggests a fundamental advancement in how LLMs can approach complex reasoning tasks that can not be replicated with prompting.

\subsection{Self-Improvement and Synthetic Data Generation}
Recent approaches to enhancing model capabilities have explored both self-improvement and synthetic data generation. The Self-Taught Reasoner (STaR) method \cite{zelikman_star_2022} pioneered the use of bootstrapping techniques, demonstrating that models can enhance their reasoning abilities by iteratively solving problems and learning from their own solutions. This concept has been extended by methods like Tree of Thoughts \cite{yao_tree_2023} and Least-to-Most \cite{zhou_least--most_2023}, which generate more structured reasoning paths.

However, these approaches have primarily focused on generating either direct solutions or simple step-by-step reasoning. Our work bridges this gap by introducing a more sophisticated pipeline that explicitly focuses on generating and validating complex problem-solving demonstrations. We combine the iterative improvement aspects of STaR with structured quality criteria to generate synthetic data that captures the full breadth of human-like problem exploration and solution refinement.

\subsection{Test Time Compute and Solution Space Exploration}

Recent work has demonstrated the significant impact of test-time compute scaling on model performance \cite{snell_scaling_2024}, particularly for complex reasoning tasks. While methods like Monte Carlo Tree Search (MCTS)\cite{xie_monte_2024} have proven effective in domains with well-defined state spaces and clear evaluation metrics (e.g., chess and Go), applying similar systematic search strategies to language model reasoning is less straightforward. This is because the solution space for mathematical or logical reasoning problems is often less structured and more difficult to systematically explore compared to game trees.

Chain-of-Thought (CoT) prompting has emerged as a more fluid alternative for scaling test-time compute in language models. Rather than attempting to implement rigid tree search algorithms, CoT allows models to naturally explore the solution space through verbalized reasoning steps. This approach can be viewed as a form of "soft" search over the solution space, where each step of the reasoning process represents an exploration of potential solution paths\cite{wang_towards_2023}. Various approaches have been proposed to enhance this exploration, from simple best-of-N sampling to more sophisticated beam search with verifier models\cite{snell_scaling_2024}.

Our work takes a different approach. Rather than focusing on test-time compute scaling through multiple solution attempts, we concentrate on enhancing the quality and depth of individual Chain-of-Thought responses. Our hypothesis is that by training models to engage in more thorough and systematic reasoning processes, we can achieve many of the benefits of test-time compute scaling without the computational overhead of generating multiple solutions. The success of our approach suggests that improving the quality of single-pass reasoning may be more efficient than relying on multiple solution attempts, though these approaches are not mutually exclusive.

\section{Method}

We first developed a high-quality dataset comprising detailed worked solutions from the American Invitational Mathematics Examination (AIME).

AIME was selected as our source material because its problems typically do not  demand advanced mathematical techniques, but they require sophisticated logical reasoning and problem-solving skills. This combination makes AIME problems particularly suitable for studying mathematical reasoning capabilities in LLMs. All human-annotated data in our dataset was sourced exclusively from AIME 2023 and before. AIME 2024 was used as our test set. Our results represent the average performance across three complete runs of all 30 questions on AIME 2024.

\begin{figure}[h]
\begin{tcolorbox}[
    colback=blue!5,
    colframe=blue!50!black,
    fonttitle=\bfseries
]
\textbf{Question:} Lilypads $1,2,3,\ldots$ lie in a row on a pond. A frog makes a sequence of jumps starting on pad $1$. From any pad $k$ the frog jumps to either pad $k+1$ or pad $k+2$ chosen randomly with probability $\tfrac{1}{2}$ and independently of other jumps. The probability that the frog visits pad $7$ is $\tfrac{p}{q}$, where $p$ and $q$ are relatively prime positive integers. Find $p+q$.

\medskip
\textbf{Intial Human Solution:} First let's understand the question. 123 lilly pads in a row on a lake. Okay, one two three lillpyads and a frog makes a sequence of jumps starting at pad one. From any pad k the frog can jump to k + one or k plus two. Ahh I just made a slight mistake it's one two three ... so there can be more than 3 lilly pads otherwise it wouldn't make sense because where would he jump for state 2? So there's basically a real integer number of pads, okay. So he starts at number one and you can jump 1 forward or 2 forward, okay. So essentially every pad you can jump over one step forward or two steps forward.

Probability visits pad 7 is P over Q relative prime integers. Find P over Q. Okay so essentially we want to see what is the probability of him reaching 7, right? So based on this this looks like some kind of like Markov chain problem where we can do some kind of backtracking recursion. That's a little complicated, is there like a simple way of solving this? Looking at it doesn't really seem like that. Initial brain strom and relating to what we know It looks like a markov chain problem. Maybe there's some way to do it with like a binomial of simpler way than just backtracking. Back tracking seems like a pretty easy method though...
\end{tcolorbox}
\label{fig:initial-workings}
\end{figure}

\subsection{Creating Worked Solutions}
\label{sec:creating-worked-solutions}

To develop high-quality worked solutions, we developed a multi-stage process combining the expertise of graduate-level mathematics students with large language model capabilities.\footnote{The first author is the annotator in our data creation process.} The process begins with these expert solvers verbalizing their complete thought process while solving mathematical problems, captured through speech-to-text technology. We found this method particularly effective at documenting authentic problem-solving processes, including crucial elements such as initial brainstorming, strategy selection, and the development of mathematical intuition.

A significant advantage of speech-to-text capture was its ability to record authentic problem-solving techniques that might be omitted in written solutions, such as trial-and-error approaches and the strategy of solving simpler cases to build intuition. However, this initial stage proved challenging, as participants often struggled to fully articulate their implicit problem-solving steps and mathematical reasoning. An example of the raw working out can be seen in Figure \ref{fig:initial-workings}.

We then experimented with experts using LLMs to transform their conceptual approaches into detailed solutions. Importantly, we made a deliberate decision to minimize human computational errors in this initial dataset. Our preliminary experiments showed that including human algebraic manipulation errors actually degraded LLM performance, causing models to introduce errors in cases where they previously produced correct calculations. This observation suggests that LLMs may learn to mimic human error patterns when exposed to them in training data, highlighting the importance of maintaining computational accuracy in the initial solution generation phase.

Concretely, the speech-to-text output undergoes initial manual cleanup to correct transcription errors and provide basic structure while preserving the natural flow of problem-solving logic. We then processed this refined solution through large language models---specifically GPT-4o mini and o1 mini---to generate detailed explanations of the initial calculations. GPT 4o mini and O1 mini were used due to their ability generate detailed verbose response \footnote{While we primarily used these two models, comparable results can be achieved with open-source alternatives, such as Llama 405B. \url{https://github.com/tamassimonds/REL/blob/main/samples/Llama405B.pdf}}.. We note that O1 mini hides it's CoT and has prevention in place to prevent you from attempting to prompt to get it's CoT. Initial testing indicated that while any language model could be utilized for this purpose, certain LLMs prove more time-efficient than others by requiring less human intervention. 


The LLM serves to expand on the human expert's work. They were instructed to do things like add more brainstorming and more reflection and checking throughout the solution while preserving the humans general solution path. 

The enhancement process follows two distinct paths based on solution correctness. For incorrect human solutions, we instructed the LLM to preserve the original flawed reasoning while adding reflective transitions (e.g., "Ah, I see I've made a mistake here") before integrating correct reasoning from verified AIME solutions. This approach creates content that demonstrates both common pitfalls and their corrections. For correct solutions lacking sufficient depth, we provided the LLM with additional example solutions and prompted it to generate additional verification steps (e.g., "Let me check another method to verify this"). Often multiple prompt chats were needed to obtain desired output. Both paths required careful manual review to seamlessly stitch together multiple LLM-generated solutions into coherent narratives that maintained authentic problem-solving flows.\footnote{A sample chat for creating the worked solution for a problem can be seen at: \url{https://github.com/tamassimonds/REL/blob/main/samples/gpt4omini_data_creation_process2.pdf}}.

In the final stage, human editors played a crucial role in enhancing and refining the AI-generated content. A key focus was incorporating more substantive reflection and verification, as we found that while language models attempted to reflect on their work, these reflections tended to be superficial rather than engaging in genuine critical analysis. Human editors addressed this by adding meaningful self-checks and thorough validation of previous answers. This final human editing stage was crucial for creating solutions that felt natural while incorporating multiple solution strategies and verification steps.

While graduate-level mathematics students could independently produce these detailed solutions, our hybrid human-LLM collaborative approach demonstrated better efficiency and effectiveness. The initial human input proved essential, as it captured authentic problem-solving approaches that LLMs consistently failed to generate on their own. Despite extensive experimentation with sophisticated prompting techniques, we found fundamental limitations in the LLMs' ability to replicate genuine mathematical problem-solving strategies. The models struggled particularly with core mathematical practices like exploring simpler cases to build intuition or developing strategic approaches to complex problems.

We initially just tried prompting the LLMs explicitly to demonstrate behaviors like reflection and backtracking without the human data, but these attempts typically resulted in superficial performances rather than genuine analytical thinking. The models would often produce what appeared to be thoughtful self-analysis but failed to actually verify their work or identify genuine errors in their reasoning \cite{huang_large_2024}. This limitation highlighted a crucial gap between mimicking the appearance of mathematical reasoning and engaging in authentic problem-solving processes. It also highlights the necessity of the initial human solutions.

\subsection{Reasoning Enhancement Loop (REL)}

We first finetune an LLM (generator) on the human-created worked solutions created in the previous section.\footnote{Despite extensive experimentation with prompting strategies, we found that LLMs could not reliably exhibit the systematic planning and exploratory working-out behaviors necessary for effective REL without first finetuning on human data.}
We focused primarily on finetuning GPT-4o and GPT-4o-mini as they provide the most direct comparison to O1 and O1-mini's performances. We used OpenAI's fine-tuning API for fine tuning.

 \begin{figure}[htbp]
    \centering
    \includegraphics[width=1\textwidth]{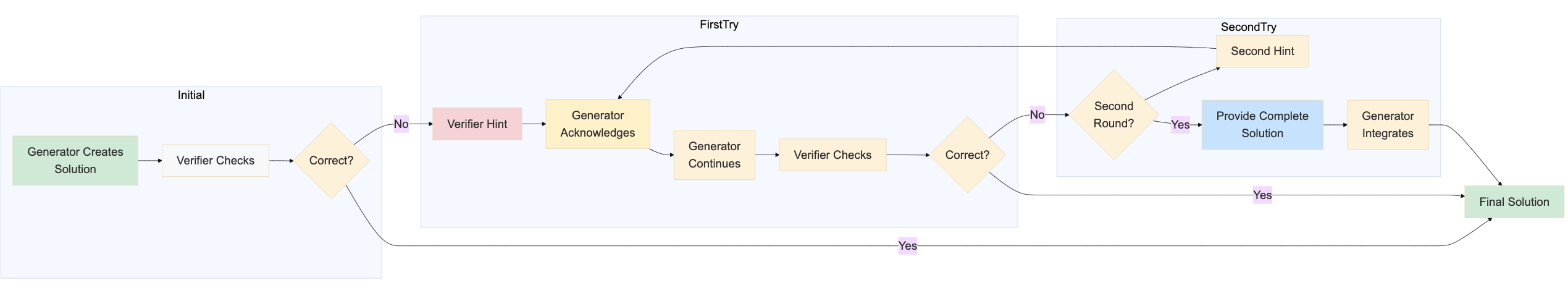}
    \caption{The Reasoning Enhancement Loop (REL) process. The system begins with initial training on human-generated solutions, then enters an iterative loop where the solution generator creates solutions that are verified and corrected through a hint-based process. Successful solutions are added to the training dataset for model refinement.}
    \label{fig:rel-process}
\label{REL_Diagram}
\end{figure}

After establishing baseline reasoning capabilities through initial finetuning, we next introduced Reasoning Enhancement Loop (REL), a self-improvement pipeline designed to generate more worked solutions for new problems. The novelty of REL is that it involves both exploratory reasoning and self-verification (mimicking natural problem-solving), where errors are discovered and corrected during the solution process itself.

REL consists of two primary components: our finetuned model serving as the solution generator, and a verification model that identifies mathematical or logical errors. Unlike traditional approaches that aim to eliminate errors entirely, REL intentionally preserves and enhances the model's ability to detect and correct its own mistakes during problem-solving(\autoref{fig:rel-process}). For verifier model we used GPT 4o prompted with the correct solution and the generators working out.

When errors are found by the verifier, instead of discarding the work and starting over, we implement an iterative hint-based correction process. The verifier first identifies where the solution went wrong and provides a targeted hint about the error. The verifier has access to the provided sample AIME solutions. The hint does not give the full solution but just a hint to the model of how to fix the issue in the solution. The generator is then passed it's previous working out and the hint and asked to rewrite to keep it's initial working out the same but to acknowledging its mistake (e.g., "Ah, I see I've made an error...") before proceeding with the correction. If the solution remains incorrect after two rounds of hints, we provide the complete correct solution and tell the generator to naturally intergrate this into into solution as if it discovered it. For example, if a geometry problem contained an incorrect angle calculation, the verifier might hint "Check your calculation of the supplementary angle in step 3," and the generator would append to its existing solution: "Ah, I see I've made an error in calculating the supplementary angle. Let me revise my calculation..." This approach maintains solution continuity while teaching the model to naturally incorporate feedback and corrections into its reasoning process.

Solutions generated through REL are incorporated into subsequent training iterations. We then finetune the generator on this new dataset and repeat the process. To summarise, for our experiments we first finetuned the model of the dataset of 100 human-created worked solutions and then a generated 100 samples for each iteration of the REL process.

Note that when training GPT-4o-mini and Llama 3.2 3B, we found it better to train using the outputs of GPT-4o from REL. We noticed that the worked solutions generated by smaller models are of poor quality, implying ultimately we still need a model of a certain capacity for generating synthetic worked solutions in REL (this also means the training of our smaller models can be interpreted as a form of knowledge distillation).


\section{Results}
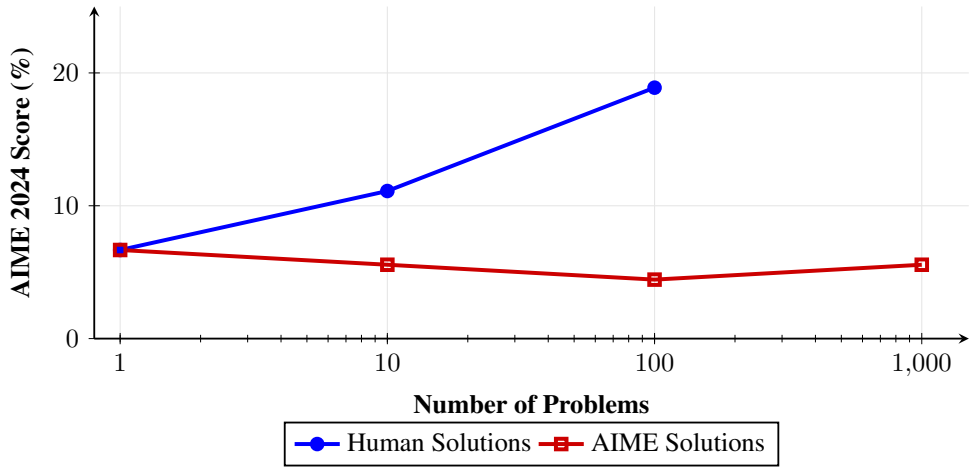
\begin{figure}[h]
\centering
\begin{tikzpicture}
\begin{semilogxaxis}[
    width=0.8\textwidth,
    height=6cm,
    xlabel={\textbf{Number of Problems}},
    ylabel={\textbf{AIME 2024 Score (\%)}},
    title={\textbf{Human vs AIME Performance}},
    legend style={
        at={(0.5,-0.25)},
        anchor=north,
        legend columns=2
    },
    grid=major,
    grid style={draw=gray!20},
    xmin=0.8,
    xmax=1500,
    ymin=0,
    ymax=25,
    xtick={1,10,100,1000},
    tick style={draw=black},
    log ticks with fixed point,
    axis lines=left,
    line width=0.7pt,
    title style={font=\large},
    label style={font=\normalsize}
]
\addplot[blue,mark=*,line width=1.5pt] coordinates {
    (1, 6.67)
    (10, 11.11)
    (100, 18.89)
};
\addplot[red!80!black,mark=square,line width=1.5pt] coordinates {
    (1, 6.67)
    (10, 5.56)
    (100, 4.44)
    (1000, 5.56)
};
\legend{Human Solutions, AIME Solutions}
\end{semilogxaxis}
\end{tikzpicture}
\caption{Performance comparison between models trained on human-generated worked solutions versus standard AIME solutions. Results show superior scaling of human-annotated solutions compared to traditional AIME solution sets.}
\label{fig:human-vs-aime}
\end{figure}

\subsection{Human Generation Solutions vs AIME Solutions}


Our initial experiments focused on comparing two approaches to fine-tuning GPT-4o mini: using human-generated worked solutions versus the provided AIME solutions. The aim is to show that any performance improvement isn't simply due to domain specialization, but that the nature of the fine-tuning data is important. For clarity, we refer to these models as Human FT GPT-4o mini (fine-tuned on human-generated solutions described in Section 3.1) and AIME FT GPT-4o mini (fine-tuned on sample solutions provided by AIME). Sample solutions provided for the AIME problems gave direct step by step to the final solution, and so are very different to the worked solutions.

We evaluated these approaches on AIME problems from 2023 and earlier; results in Figure \ref{fig:human-vs-aime}. For these results, we vary the number of fine-tuning data (x-axis), from 1 sample to 1K samples. Note that we only have results for Human FT GPT-4o mini for up to 100 samples---that's because that's the maximum number of worked solutions we have created in Section \ref{sec:creating-worked-solutions}.

GPT-4o mini, the base model without any fine-tuning, had accuracy of 6.66\%. AIME FT GPT-4o mini's performance remained relatively flat (4-6\%) (Figure \ref{fig:human-vs-aime}) regardless of dataset size. In contrast, Human FT GPT-4o mini demonstrated consistent improvement, reaching 18.89\% accuracy with 100 training examples—a 12.2\% increase over the baseline. The scaling behavior was particularly notable: Human FT GPT-4o mini showed a 7.8\% accuracy improvement when increasing from 10 to 100 training examples, while AIME FT GPT-4o mini exhibited minimal scaling benefits. These results suggest that the additional problem-solving steps captured in human annotations provide more effective learning signals compared to standard solutions alone.

Our qualitative analysis revealed distinct solution patterns emerging from different training approaches. Models trained with human-annotated solutions exhibited a clear developmental progression: initial training (1-10 examples) produced more verbose solutions with detailed intermediate steps, while more sophisticated behaviors—including brainstorming and solution backtracking—emerged only after extended training (50-100 examples). This progression was reflected quantitatively, with Human FT GPT-4o mini generating solutions averaging 2,152 tokens compared to the baseline's 826 tokens.

AIME FT GPT-4o mini showed an inverse pattern: as training data increased, solutions became increasingly terse, mirroring the concise style of standard AIME solutions. This in turn led to the model showing less chain of thought, which we speculate was the reason for the performance reduction.

The stark contrast in results—with Human FT GPT-4o mini significantly outperforming the AIME-specialized model—confirms that the improvements stem from the quality and structure of our human-created training data rather than mere exposure to AIME problems. This finding highlights three crucial insights: first, the fundamental importance of thorough chain-of-thought demonstrations in training data; second, the value of carefully curated human-generated solutions in developing mathematical reasoning capabilities; and third, the data efficiency of this approach, as models began exhibiting some of o1's distinctive reasoning patterns after exposure to just 50 human-generated examples, demonstrating that high-quality human demonstrations can drive significant improvements even with relatively small datasets.


\subsection{REL Process Evaluation}
\begin{figure}[!ht] 
\centering
\begin{minipage}{0.48\textwidth}
\begin{tikzpicture}
\begin{axis}[
    width=\textwidth,
    height=7cm,
    xlabel={\textbf{Training Stage}},
    ylabel={\textbf{AIME Score (\%)}},
    title={\textbf{REL Iteration Performance}},
    grid=major,
    grid style={line width=0.2pt, draw=gray!30},
    xmin=-0.2,
    xmax=4.2,
    ymin=0,
    ymax=45,
    xtick={0,1,2,3,4},
    xticklabels={Base, Human Data, Iter 1, Iter 2, Iter 3},
    x tick label style={rotate=45, anchor=east},
    tick style={draw=black},
    axis lines=left,
    line width=0.8pt,
    title style={font=\large},
    label style={font=\normalsize}
]
\addplot[
    blue,
    mark=*,
    line width=1.2pt,
    mark size=3pt
] coordinates {
    (0, 12)
    (1, 22.22)
    (2, 25.56)
    (3, 25.56)
    (4, 27.78)
};
\end{axis}
\end{tikzpicture}
\end{minipage}
\hfill
\begin{minipage}{0.48\textwidth}
\begin{tikzpicture}
\begin{axis}[
    width=\textwidth,
    height=7cm,
    xlabel={\textbf{Model}},
    ylabel={\textbf{AIME Score (\%)}},
    title={\textbf{Model Performance Comparison}},
    grid=major,
    grid style={line width=0.2pt, draw=gray!30},
    xmin=0.5,
    xmax=3.5,
    ymin=0,
    ymax=45,
    xtick={1,2,3},
    xticklabels={GPT-4o, REL FT, O1},
    x tick label style={rotate=45, anchor=east},
    tick style={draw=black},
    axis lines=left,
    line width=0.8pt,
    title style={font=\large},
    label style={font=\normalsize},
    ybar,
    bar width=25pt,
    nodes near coords,
    nodes near coords align={above}
]
\addplot[
    fill=cyan,
    draw=cyan,
] coordinates {
    (1, 12.0)
    (2, 27.78)
    (3, 44.6)
};
\end{axis}
\end{tikzpicture}
\end{minipage}
\caption{Left: Performance improvement across REL iterations. Right: Final performance comparison between base GPT-4o, our REL fine-tuned GPT-4o, and O1 on AIME 2024.}
\label{fig:rel-model-comparison}
\end{figure}
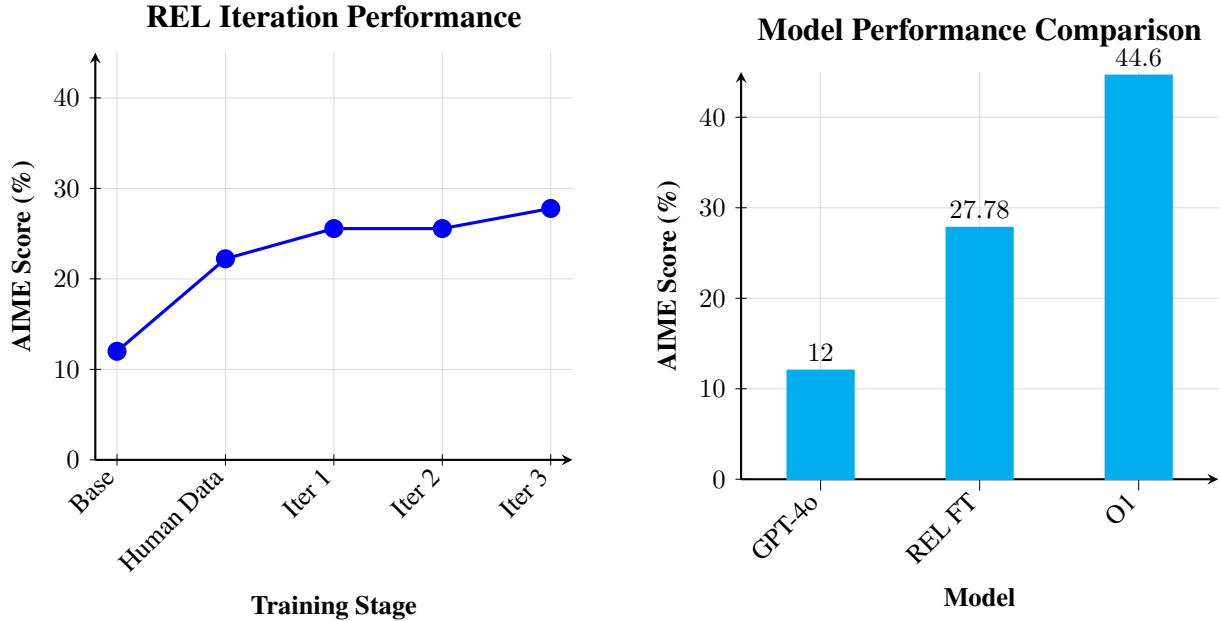



We investigate whether iterative refinement through our Reasoning Enhancement Loop (REL) can further improve performance beyond the gains achieved with human-annotated solutions alone. Our analysis examines the performance progression of GPT-4o through multiple iterations of the REL process, comparing the final REL-enhanced model (REL GPT-4o) against GPT-4o fine-tuned using only human-created worked solutions (Human GPT-4o) and OpenAI's O1 model.
Results are shown in Figure \ref{fig:rel-model-comparison}, with the iterative progression displayed in the left plot and final model comparisons in the right plot.

The left plot reveals the step-wise improvement through the REL process. Starting from GPT 4o baseline accuracy of 12.0\%, the initial introduction of human-annotated training data yielded the most substantial gain, reaching 22.22\%. Subsequent REL iterations showed more modest improvements, with the third iteration achieving 27.8\% accuracy. The right plot 
contextualizes these gains against current state-of-the-art models: while REL GPT-4o (27.78\%) more than doubled the performance of the base GPT-4o (12.0\%), it still trails behind O1's performance (44.6\%) on AIME 2024 problems. These results demonstrate that our REL process can effectively build upon the initial benefits of human-annotated data to further enhance mathematical reasoning capabilities.

We analyze the qualitative similarities in problem-solving approaches between our REL GPT-4o and OpenAI's O1 model. This comparison examines whether our REL process induces the distinctive reasoning patterns observed in more advanced models.
A comparison of solution approaches between these models is shown in \autoref{fig:solution-comparison}.
Direct examination reveals several shared problem-solving patterns between REL GPT-4o and O1. Both models consistently demonstrate: (1) explicit brainstorming at problem initiation, (2) systematic enumeration of solution strategies, and (3) integrated self-correction mechanisms. A notable example occurs in their polynomial transformation approaches, where both models exhibit mid-solution strategy revision. We also observe shared metacognitive behaviors, with both models actively monitoring their solution progress and adjusting their approaches upon discovering new insights or potential issues.

Qualitatively, we observed in the model outputs that the REL process showed distinct strengths in enhancing solution verbosity and introducing additional reflection steps, with models increasingly demonstrating thorough Chain-of-Thought reasoning and solution verification across iterations. However, our analysis revealed that fundamental problem-solving techniques were primarily determined by the initial human demonstrations. For instance, strategies like solving a simpler version of the problem (e.g., splitting "rollercoaster" into "roller" + "coaster") only emerged when explicitly shown in human examples. Similarly, while trial-and-error approaches proved effective for simpler problems, models would only attempt this strategy if it appeared in the initial training data. While REL effectively amplified these established techniques through more detailed working out and frequent self-correction, it did not independently discover more novel problem-solving strategies. This finding highlights the critical role of high-quality human demonstrations in establishing core reasoning approaches, which REL can then enhance but not fundamentally alter.

Despite REL's effectiveness in improving mathematical reasoning, we observed several issues. Models occasionally produced unnecessarily verbose solutions for simple problems and showed signs of collapse even with our relatively small 3-million-token training corpus. Some instances resulted in repetitive self-correction loops where models would repeatedly revise their work without making meaningful progress. While these issues did not significantly impact overall performance, they highlight potential instabilities in iterative refinement processes.

The strong performance gains observed through our limited human-created worked solutions and REL iterations suggest promising potential for approaching O1's benchmark 44.6\% AIME accuracy. Our results suggest with additional human data and further REL iterations we could substantially close this performance gap.


\begin{figure}[H]
\begin{tcolorbox}[
    colback=gray!5,
    colframe=gray!50!black,
    title=\textbf{Problem Statement},
    fonttitle=\bfseries
]
Let $n$ be an even positive integer. Let $p$ be a monic, real polynomial of degree $2n$; that is to say,
$p(x) = x^{2n} + a_{2n-1}x^{2n-1} + \cdots + a_1x + a_0$
for some real coefficients $a_0, \ldots, a_{2n-1}$. Suppose that
$p(\frac{1}{k}) = k^2$
for all integers $k$ such that
$1 \leq |k| \leq n$. Find all other real numbers $x$ for which
$p(\frac{1}{x}) = x^2$.
\end{tcolorbox}

\vspace{0.5cm}

\begin{minipage}[t]{0.48\textwidth}
\begin{tcolorbox}[
    colback=blue!5,
    colframe=blue!50!black,
    title=\textbf{REL FT GPT 4o},
    fonttitle=\bfseries,
    height=18cm
]
\textbf{First Thoughts:}
\begin{itemize}
\item \textbf{Symmetry?} The condition involves reciprocals and squares, which suggests symmetry in the polynomial.
\item \textbf{Roots Idea:} Maybe if $p\left(\frac{1}{k}\right) = k^2$, then $\frac{1}{k}$ are roots of some related equation.
\item \textbf{Polynomial Transformation:} Perhaps transforming the polynomial could help.
\item \textbf{Using Vieta's Formulas:} Since it's monic, the sum and product of roots might play a role.
\end{itemize}

\textbf{Possible Strategies:}
\begin{enumerate}
\item \textbf{Create a New Polynomial:} Define a new polynomial based on the given condition and see what its roots are.
\item \textbf{Leverage Symmetry:} Exploit any symmetrical properties in the polynomial.
\item \textbf{Factor the Polynomial:} Try to factor $p(x)$ using the given conditions.
\item \textbf{Use Polynomial Identities:} Apply known identities or theorems related to reciprocals.
\end{enumerate}

\vspace{0.5cm}
\textcolor{gray}{\hrulefill\ ...Working Out...\ \hrulefill}
\vspace{0.5cm}

Wait a minute, $p\left(\frac{1}{x}\right)$ is in terms of $\frac{1}{x}$, so $q(x)$ might have a higher degree when expressed in terms of $x$.

\textbf{Need to Re-express $q(x)$:} Let's re-express $q(x)$ as a polynomial in $x$.

\textbf{New Approach:}\\
Wait a Second: Since $n$ is even and $n \geq 2$, all $k$ from $1$ to $n$ and their negatives are included. So, $x = \pm 1$ are already accounted for...
\end{tcolorbox}
\end{minipage}
\hfill
\begin{minipage}[t]{0.48\textwidth}
\begin{tcolorbox}[
    colback=green!5,
    colframe=green!50!black,
    title=\textbf{O1},
    fonttitle=\bfseries,
    height=18cm
]
In other words, we are to find all real numbers $x$ such that $p(1/x)=x^2$.

Firstly, note that:
\begin{itemize}
\item The monic polynomial $p(x)$ is of degree $2n$.
\item We have $2n$ data points: for $k=\pm1$ to $\pm n$, $p(1/k)=k^2$.
\end{itemize}

Thus, we have $2n$ distinct values of $\frac{1}{k}$ (since $k$ ranges over integers $\pm1$ to $\pm n$).
Therefore, we have $2n$ points $(x,y)$ of the form $(\frac{1}{k},k^2)$.

Thus, $p$ is a polynomial of degree $2n$ that satisfies $p(\frac{1}{k})=k^2$ for $2n$ values of $\frac{1}{k}$.

\vspace{0.5cm}
\textcolor{gray}{\hrulefill\ ...Working Out...\ \hrulefill}
\vspace{0.5cm}

Wait, but $p(1/x)-x^2=0$.\\
Our equation is $p(1/x)-x^2=0$.\\
Let me define $q(x)=p(1/x)-x^2$.

\vspace{0.5cm}
\textcolor{gray}{\hrulefill\ ...Working Out...\ \hrulefill}
\vspace{0.5cm}

Wait, but $x^{2n}p(1/x)$ is the "reciprocal polynomial" of $p(x)$, evaluated at $x$.\\
But since $p(x)$ is monic of degree $2n$, its reciprocal polynomial is monic of degree $2n$ as well.

Thus, let $t(x)=x^{2n}p(1/x)=P^*(x)$...
\end{tcolorbox}
\end{minipage}

\vspace{0.3cm}
\begin{tcolorbox}[
    colback=gray!5,
    colframe=gray!50!black,
    title=\textbf{Note},
    fonttitle=\bfseries
]
This comparison shows only the initial approach phase of each model's solution. Full outputs continued for several additional pages with detailed mathematical derivations and verification steps.
\end{tcolorbox}

\caption{Comparison of solution approaches between FT GPT-4o and O1 on a complex polynomial problem.}
\label{fig:solution-comparison}
\end{figure}

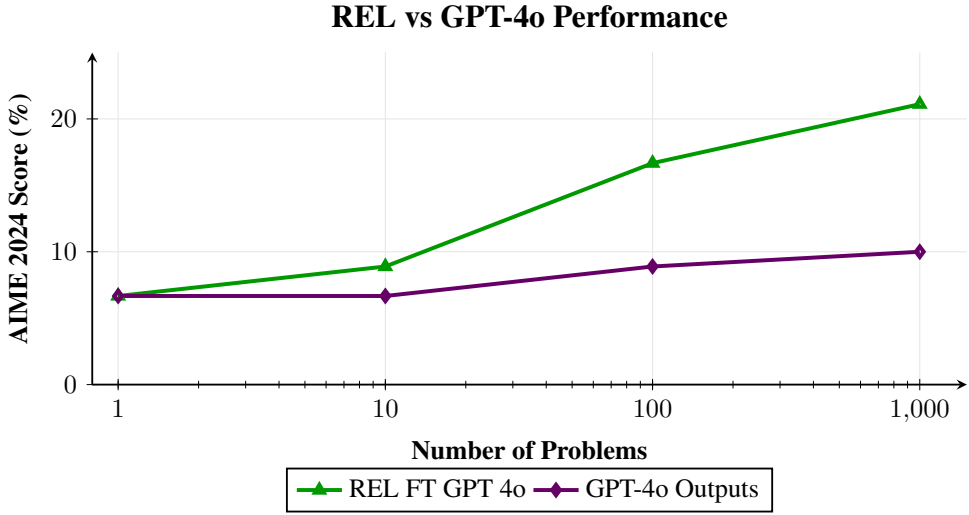
\begin{figure}[h]
\centering
\begin{tikzpicture}
\begin{semilogxaxis}[
    width=0.8\textwidth,
    height=6cm,
    xlabel={\textbf{Number of Problems}},
    ylabel={\textbf{AIME 2024 Score (\%)}},
    title={\textbf{REL vs GPT-4o Performance}},
    legend style={
        at={(0.5,-0.25)},
        anchor=north,
        legend columns=2
    },
    grid=major,
    grid style={draw=gray!20},
    xmin=0.8,
    xmax=1500,
    ymin=0,
    ymax=25,
    xtick={1,10,100,1000},
    tick style={draw=black},
    log ticks with fixed point,
    axis lines=left,
    line width=0.7pt,
    title style={font=\large},
    label style={font=\normalsize}
]
\addplot[green!60!black,mark=triangle,line width=1.5pt] coordinates {
    (1, 6.67)
    (10, 8.89)
    (100, 16.67)
    (1000, 21.11)
};
\addplot[violet!80!black,mark=diamond,line width=1.5pt] coordinates {
    (1, 6.67)
    (10, 6.67)
    (100, 8.89)
    (1000, 10.00)
};
\legend{REL FT GPT 4o, GPT-4o Outputs}
\end{semilogxaxis}
\end{tikzpicture}
\caption{Performance comparison between REL FT GPT-4o and standard GPT-4o outputs. Results demonstrate superior scaling of REL-generated solutions compared to traditional synthetic data generation approaches.}
\label{fig:rel-vs-gpt4o}
\end{figure}

\subsection{Synthetic Data Comparison}
We investigate whether synthetic data generated through our REL process can match or exceed the effectiveness of traditional synthetic data generation approaches. Our analysis compares two methods of creating training examples using GPT-4o. We compare using GPT 4o with rejection sampling and rationalization (if answer incorrect given correct answer and told to rationalize) to using data generated from our REL GPT 4o using the REL process. Both methods are then used to fine-tune GPT-4o mini, resulting in two models for comparison: REL data finteuned (REL\_Data\_FT\_mini) and rejection sampling data finetuned (RS\_Data\_FT\_mini).

Results of this comparison are shown in \autoref{fig:rel-vs-gpt4o}.
REL\_Data\_FT\_mini demonstrated superior scaling properties, achieving 21.11\% accuracy with 1000 training examples. In contrast, Baseline RS\_Data\_FT\_mini showed significantly slower improvement despite equivalent dataset sizes. The performance gap became particularly pronounced as dataset size increased, with REL-generated data maintaining consistent improvement while baseline-generated data showed diminishing returns.

Qualitative analysis of solution patterns reveals distinct behavioral differences between the models. REL\_Data\_FT\_mini consistently produced detailed, structured solutions with clear progression through problem-solving stages. These solutions typically began with explicit strategy consideration, followed by systematic exploration of solution paths, and included self-correction when necessary. In contrast, Baseline RS\_Data\_FT\_mini tended to generate more linear solutions that, while technically correct, lacked the metacognitive elements and strategic depth observed in REL-generated solutions.

These results reveal broader insights about effective methods for generating synthetic training data for language models. Both synthetic data approaches—REL-generated and baseline-generated—significantly outperformed training on raw AIME solutions, demonstrating the value of detailed solution rationales in training data. However, the rapid emergence of sophisticated reasoning patterns in models trained on just 100 REL-generated examples, compared to their slower development in baseline-generated examples, suggests that our REL process provides a more efficient method for synthetic data generation. By encoding complex problem-solving strategies more effectively, REL offers a promising approach for creating high-quality training data that can efficiently improve mathematical reasoning capabilities in language models. Additionally, this approach demonstrates a more effective method for leveraging our limited corpus of high-quality mathematical problems and solutions. By using existing question-answer pairs as seeds for generating rich, pedagogically valuable training data, REL provides a scalable way to amplify the impact of rare but valuable mathematical content.



\begin{figure}[H]
\begin{tcolorbox}[
    colback=gray!5,
    colframe=gray!50!black,
    title=\textbf{Problem Statement},
    fonttitle=\bfseries
]
\textbf{Prompt:} How many R's are in 'strawberry'?
\end{tcolorbox}

\vspace{0.5cm}

\begin{minipage}[t]{0.48\textwidth}
\begin{tcolorbox}[
    colback=blue!5,
    colframe=blue!50!black,
    title=\textbf{O1-Llama 3.2 3B Output},
    fonttitle=\bfseries,
    height=18cm
]
\textbf{First Thoughts}\\
Okay, so the question is: How many R's are in "strawberry"?

\textbf{Brainstorming:}
\begin{itemize}
\item Option 1: I could write out the word and count each letter individually.
\item Option 2: Perhaps there's a pattern or rule in English words regarding the placement of R's.
\item Option 3: Maybe I can use a dictionary or online tool to verify.
\end{itemize}

\textbf{Step 1:} Writing Out the Word\\
Let me start by writing out the word "strawberry" to see it clearly:\\
S-T-R-A-W-B-E-R-R-Y

\textbf{Step 2:} Counting the R's\\
Now, I'll focus specifically on the R's:\\
R's: R-R-Y
\begin{itemize}
\item Count: There are 2 R's in the word "strawberry".
\item Check: Let me double-check... S-T-R-A-W-B-E-R-R-Y—yes, I see two R's.
\end{itemize}

\textbf{Step 3:} Final Check
\begin{itemize}
\item Quick Check: Let me glance back at the word: S-T-R-A-W-B-E-R-R-Y
\item Final Answer: There are 2 R's in the word "strawberry".
\end{itemize}
\end{tcolorbox}
\end{minipage}
\hfill
\begin{minipage}[t]{0.48\textwidth}
\begin{tcolorbox}[
    colback=green!5,
    colframe=green!50!black,
    title=\textbf{Llama 3.1 3B Output},
    fonttitle=\bfseries,
    height=18cm
]
There are 2 R's in Strawberry
\end{tcolorbox}
\end{minipage}

\vspace{0.3cm}
\begin{tcolorbox}[
    colback=gray!5,
    colframe=gray!50!black,
    title=\textbf{Analysis},
    fonttitle=\bfseries
]
O1-Llama 3.2 3B demonstrates extensive step-by-step reasoning and self-verification, though both models arrive at the same answer. Note O1-Llama 3.2 3B's multiple verification steps and consideration of alternative approaches, showing significantly more sophisticated reasoning even for this simple task.
\end{tcolorbox}

\caption{Comparison of counting approaches between O1-Llama 3.2 3B and Llama 3.1 3B on a simple letter counting task.}
\label{fig:strawberry-comparison}
\end{figure}

\section{Discussion}
Our findings suggest that small amounts of expertly crafted data can be surprisingly effective at unlocking latent model capabilities, challenging the conventional focus on expanding model scale and pretraining data size. The dramatic performance difference between models trained on thoroughly worked solutions versus larger sets of simple solutions (10x size advantage) demonstrates that quality of reasoning demonstrations matters more than quantity.

While our REL methodology enhanced reasoning patterns, the fundamental problem-solving strategies traced back to human demonstrations, suggesting that advancing AI capabilities may depend more on effectively capturing human expertise than on scaling data collection. This points to a future direction focused on creating comprehensive libraries of expert human reasoning processes across different fields, where the key challenge lies in surfacing and refining pretrained models' latent abilities through exposure to sophisticated human problem-solving examples.

The effectiveness of targeted post-training with a limited dataset aligns with recent findings on pretraining and post-training in language models (\cite{brown_language_2020}; \cite{chowdhery_palm_2022}). Our results with 100 human demonstrations build on previous small-dataset successes (\cite{zhou_instruction-following_2023}; \cite{shumailov_curse_2024}) and show that specific capabilities can be enhanced through focused interventions (\cite{askell_general_2021}, \cite{ouyang_training_2022}). This resource-efficient approach (\cite{zhang_instruction_2024}) suggests opportunities for making AI development more accessible, with future work focused on collecting more diverse human demonstrations of expert reasoning.


The initial success of our REL methodology with AIME problems led us to develop ReasonSet, a comprehensive dataset of 2,000 question-solution pairs spanning multiple reasoning domains. Using our REL FT GPT-4o model, we generated synthetic data encompassing AIME, GPQA, MATH dataset problems, and novel general reasoning questions derived from our prompt template (Appendix A). Notably, this expansion required no additional human demonstrations beyond the original AIME examples. We are releasing ReasonSet as an open-source resource to help the broader community develop reasoning capabilities in language models.
These experiments revealed important insights about the transferability of reasoning capabilities across model scales. When fine-tuning Llama 3.2 3B on ReasonSet, we observed interesting patterns in the model's adoption of reasoning behaviors. While the model successfully incorporated surface-level planning elements like explicit brainstorming, strategy enumeration, and verification attempts (\autoref{fig:strawberry-comparison}), it struggled with deeper reasoning tasks. The significant performance gap between O1-Llama 3.2 3B and larger models like GPT-4o and GPT-4o mini suggests that although smaller models can learn to emulate sophisticated reasoning patterns, they may lack the computational capacity to fully execute complex logical reasoning chains. Nevertheless, our open-sourced O1-Llama 3.2 3B model demonstrates how readily structured reasoning behaviors can be induced through targeted fine-tuning, even in relatively small language models.

The success of our approach raises broader implications for language model training methodologies. While significant research effort has focused on developing increasingly sophisticated training techniques and architectures, our results suggest that certain advanced behaviors might be more readily obtained simply through exposure to high-quality human demonstrations. The apparent absence of sophisticated reasoning capabilities in standard language models may stem not from architectural limitations, but from the nature of their training data—which typically consists of final solutions and conclusions rather than detailed problem-solving processes. Just as our models learned to mimic human problem-solving processes from detailed mathematical demonstrations, other advanced capabilities might emerge more naturally if models were trained on data that better captures human expert thinking. This insight may extend beyond mathematical reasoning to other domains, including agent behavior and decision-making, where current training data similarly lacks explicit documentation of human decision-making processes. Rather than developing elaborate training techniques, the key challenge may be collecting high-quality demonstrations that make expert thinking processes explicit. Our findings suggest that addressing this gap in training data composition could be as impactful as architectural innovations in advancing AI capabilities.




\section{Conclusion}

In this paper, we presented several contributions to enhance reasoning capabilities in language models. First, we introduced a novel data creation methodology combining human expertise with AI assistance to efficiently generate high-quality worked solutions. This hybrid approach proved crucial for capturing authentic problem-solving processes while maintaining scalability and data quality.

Second, we developed the Reasoning Enhancement Loop (REL), a systematic pipeline for generating additional high-quality worked solutions through iterative refinement and validation. Our empirical results demonstrated that REL successfully enhanced model performance on AIME problems, achieving a 27.78\% accuracy compared to the 12\% baseline of GPT-4o, while exhibiting many of the sophisticated reasoning behaviors characteristic of O1.

Third, our experimental findings challenge conventional wisdom about scaling AI capabilities. Models trained on just 100 human-annotated worked solutions (18.89\%) significantly outperformed those trained on 1000 standard solutions (5.56\%), suggesting that the quality and structure of reasoning demonstrations matter far more than quantity. This insight points to an alternative path for advancing AI capabilities through careful curation of expert problem-solving processes rather than simply scaling up data collection.

Finally, we release two key resources to the research community: ReasonSet, a comprehensive dataset of 2,000 worked solutions across multiple reasoning domains, and O1-Llama 3.2 3B, an open-source model demonstrating that sophisticated reasoning behaviors can be induced in smaller models through targeted fine-tuning. While O1-Llama shows limitations compared to larger models, it serves as a proof-of-concept for democratizing access to advanced reasoning capabilities.

These contributions collectively suggest a promising direction for future AI development: focusing on capturing and transferring human expertise through carefully structured demonstrations rather than relying solely on increased model scale or dataset size. Our success with relatively modest computational resources indicates that significant advances in AI reasoning capabilities may be achievable through more efficient, targeted approaches to model development.


\section*{Appendix A}
\subsection*{A.1 Human Dataset Collection}

Our data collection process started with human-generated seed solutions that were then enhanced through a structured prompting methodology. The method was not completely deterministic with human hand crafted prompts often used to elicit different behaviors depending on the models response. The process used the following core prompt though:

Initial prompt with Human Solution and Questions
\begin{quote}
Rewrite the human solution showing your full working out

To get full marks you must show your thought process as well. First show you exploring a range of different strategies. If one doesn't work show it and then use another method. Brainstorm at first how to solve it. I want you to write it like a human as in start off by brainstorming. Reflect often if it's correct, check often if your approach is correct and if not change your approach. After every few steps naturally check like a human would if the answer is correct before continuing.

First thoughts are important - write down what initially comes to mind, even if it seems messy or incomplete. Like when you first see a problem and think 'This looks similar to something I've done before...' or 'I have no idea where to start...'

Let your thinking flow naturally:
\begin{itemize}
\item If you suddenly realize something isn't working: "Wait, that can't be right because..."
\item When you get stuck: "Hmm, maybe I need to back up and try..."
\item When something clicks: "Oh! This reminds me of..."
\end{itemize}

Check your work like you naturally would:
\begin{itemize}
\item Sometimes quick checks: "That number seems too big..."
\item Sometimes deeper reflection: "Let me think about whether this approach makes sense..."
\item Random insights: "Actually, there might be an easier way..."
\end{itemize}

Don't be afraid to:
\begin{itemize}
\item Show your mistakes and dead ends
\item Change direction when something feels off
\item Think out loud about your confusion or uncertainty
\item Have "aha moments" in the middle of working
\end{itemize}
\end{quote}

For incorrect solutions, we used the following continuation prompt. We also provide the provided AIME solution:
\begin{quote}
Correct your last one. Write something like "Ah, I see I've made an error." Write it out like a continuation from your previous incorrect solution like you are a human. Show your process of discovering the last solution is wrong and show your working out in leading to new correct solution. Don't acknowledge you have been given the solution. Naturally discover the mistake you made in your last one. If there are multiple solutions incorporate more than one to validate your solution.

To get full marks you must show your thought process aswell. First show you exploring a range of different stratergies. If one doesn't work show it and then use another method. Brainstorm at first how to solve it. I want you to write it like a human as in start off by brainstorming. Reflect often if it's correct, check often if your approach is corret and if not change your approach. After every few steps naturally check like a human would if the answer is correct before continueing 
 
\end{quote}

For solutions that were correct but needed more detail we provide the AIME sample solution and prompt it with the following:
\begin{quote}
Here are some other solutions. Write them out like a continuation of your last answer as a way to further validate your solution and to incorporate more brainstorming.

To get full marks you must show your thought process aswell. First show you exploring a range of different stratergies. If one doesn't work show it and then use another method. Brainstorm at first how to solve it. I want you to write it like a human as in start off by brainstorming. Reflect often if it's correct, check often if your approach is corret and if not change your approach. After every few steps naturally check like a human would if the answer is correct before continueing 
\end{quote}

We often had to prompt the model with the following to get more details
\begin{quote}
    To get full marks you must show your thought process aswell. First show you exploring a range of different stratergies. If one doesn't work show it and then use another method. Brainstorm at first how to solve it. I want you to write it like a human as in start off by brainstorming. Reflect often if it's correct, check often if your approach is corret and if not change your approach. After every few steps naturally check like a human would if the answer is correct before continueing 
    Show alot more brainstorming and exploration just like a human
\end{quote}

Again we restate that prompting was not an automatic process and often needed a lot of back and forward between LLM and humans. Often humans had to prompt to give more detail in explict points or reflect on certain other points. 


\begin{thebibliography}{10}

\bibitem{askell_general_2021}
Amanda Askell, Yuntao Bai, Anna Chen, Dawn Drain, Deep Ganguli, Tom Henighan, Andy Jones, Nicholas Joseph, Ben Mann, Nova DasSarma, Nelson Elhage, Zac Hatfield-Dodds, Danny Hernandez, Jackson Kernion, Kamal Ndousse, Catherine Olsson, Dario Amodei, Tom Brown, Jack Clark, Sam McCandlish, Chris Olah, and Jared Kaplan.
\newblock A {General} {Language} {Assistant} as a {Laboratory} for {Alignment}, December 2021.
\newblock arXiv:2112.00861.

\bibitem{brown_language_2020}
Tom~B. Brown, Benjamin Mann, Nick Ryder, Melanie Subbiah, Jared Kaplan, Prafulla Dhariwal, Arvind Neelakantan, Pranav Shyam, Girish Sastry, Amanda Askell, Sandhini Agarwal, Ariel Herbert-Voss, Gretchen Krueger, Tom Henighan, Rewon Child, Aditya Ramesh, Daniel~M. Ziegler, Jeffrey Wu, Clemens Winter, Christopher Hesse, Mark Chen, Eric Sigler, Mateusz Litwin, Scott Gray, Benjamin Chess, Jack Clark, Christopher Berner, Sam McCandlish, Alec Radford, Ilya Sutskever, and Dario Amodei.
\newblock Language {Models} are {Few}-{Shot} {Learners}, July 2020.
\newblock arXiv:2005.14165.

\bibitem{chowdhery_palm_2022}
Aakanksha Chowdhery, Sharan Narang, Jacob Devlin, Maarten Bosma, Gaurav Mishra, Adam Roberts, Paul Barham, Hyung~Won Chung, Charles Sutton, Sebastian Gehrmann, Parker Schuh, Kensen Shi, Sasha Tsvyashchenko, Joshua Maynez, Abhishek Rao, Parker Barnes, Yi~Tay, Noam Shazeer, Vinodkumar Prabhakaran, Emily Reif, Nan Du, Ben Hutchinson, Reiner Pope, James Bradbury, Jacob Austin, Michael Isard, Guy Gur-Ari, Pengcheng Yin, Toju Duke, Anselm Levskaya, Sanjay Ghemawat, Sunipa Dev, Henryk Michalewski, Xavier Garcia, Vedant Misra, Kevin Robinson, Liam Fedus, Denny Zhou, Daphne Ippolito, David Luan, Hyeontaek Lim, Barret Zoph, Alexander Spiridonov, Ryan Sepassi, David Dohan, Shivani Agrawal, Mark Omernick, Andrew~M. Dai, Thanumalayan~Sankaranarayana Pillai, Marie Pellat, Aitor Lewkowycz, Erica Moreira, Rewon Child, Oleksandr Polozov, Katherine Lee, Zongwei Zhou, Xuezhi Wang, Brennan Saeta, Mark Diaz, Orhan Firat, Michele Catasta, Jason Wei, Kathy Meier-Hellstern, Douglas Eck, Jeff Dean, Slav Petrov, and Noah Fiedel.
\newblock {PaLM}: {Scaling} {Language} {Modeling} with {Pathways}, October 2022.
\newblock arXiv:2204.02311.

\bibitem{huang_large_2024}
Jie Huang, Xinyun Chen, Swaroop Mishra, Huaixiu~Steven Zheng, Adams~Wei Yu, Xinying Song, and Denny Zhou.
\newblock Large {Language} {Models} {Cannot} {Self}-{Correct} {Reasoning} {Yet}, March 2024.
\newblock arXiv:2310.01798.

\bibitem{kambhampati_llms_2024}
Subbarao Kambhampati, Karthik Valmeekam, Lin Guan, Mudit Verma, Kaya Stechly, Siddhant Bhambri, Lucas Saldyt, and Anil Murthy.
\newblock {LLMs} {Can}'t {Plan}, {But} {Can} {Help} {Planning} in {LLM}-{Modulo} {Frameworks}, June 2024.
\newblock arXiv:2402.01817.

\bibitem{gpt4osystemcard}
OpenAI.
\newblock Open{AI} {GPT4o} system card.
\newblock {\em arXiv preprint arXiv:1804.09028}, 2024.

\bibitem{o1systemcard}
OpenAI.
\newblock Open{AI} o1 system card.
\newblock {\em arXiv preprint arXiv:1804.09028}, 2024.

\bibitem{ouyang_training_2022}
Long Ouyang, Jeff Wu, Xu~Jiang, Diogo Almeida, Carroll~L. Wainwright, Pamela Mishkin, Chong Zhang, Sandhini Agarwal, Katarina Slama, Alex Ray, John Schulman, Jacob Hilton, Fraser Kelton, Luke Miller, Maddie Simens, Amanda Askell, Peter Welinder, Paul Christiano, Jan Leike, and Ryan Lowe.
\newblock Training language models to follow instructions with human feedback, March 2022.
\newblock arXiv:2203.02155.

\bibitem{shumailov_curse_2024}
Ilia Shumailov, Zakhar Shumaylov, Yiren Zhao, Yarin Gal, Nicolas Papernot, and Ross Anderson.
\newblock The {Curse} of {Recursion}: {Training} on {Generated} {Data} {Makes} {Models} {Forget}, April 2024.
\newblock arXiv:2305.17493.

\bibitem{snell_scaling_2024}
Charlie Snell, Jaehoon Lee, Kelvin Xu, and Aviral Kumar.
\newblock Scaling {LLM} {Test}-{Time} {Compute} {Optimally} can be {More} {Effective} than {Scaling} {Model} {Parameters}, August 2024.
\newblock arXiv:2408.03314.

\bibitem{valmeekam_llms_2024}
Karthik Valmeekam, Kaya Stechly, and Subbarao Kambhampati.
\newblock {LLMs} {Still} {Can}'t {Plan}; {Can} {LRMs}? {A} {Preliminary} {Evaluation} of {OpenAI}'s o1 on {PlanBench}, September 2024.
\newblock arXiv:2409.13373.

\bibitem{wang_towards_2023}
Boshi Wang, Sewon Min, Xiang Deng, Jiaming Shen, You Wu, Luke Zettlemoyer, and Huan Sun.
\newblock Towards {Understanding} {Chain}-of-{Thought} {Prompting}: {An} {Empirical} {Study} of {What} {Matters}, June 2023.
\newblock arXiv:2212.10001.

\bibitem{wei_chain--thought_2023}
Jason Wei, Xuezhi Wang, Dale Schuurmans, Maarten Bosma, Brian Ichter, Fei Xia, Ed~Chi, Quoc Le, and Denny Zhou.
\newblock Chain-of-{Thought} {Prompting} {Elicits} {Reasoning} in {Large} {Language} {Models}, January 2023.
\newblock arXiv:2201.11903.

\bibitem{xie_monte_2024}
Yuxi Xie, Anirudh Goyal, Wenyue Zheng, Min-Yen Kan, Timothy~P. Lillicrap, Kenji Kawaguchi, and Michael Shieh.
\newblock Monte {Carlo} {Tree} {Search} {Boosts} {Reasoning} via {Iterative} {Preference} {Learning}, June 2024.
\newblock arXiv:2405.00451.

\bibitem{yao_tree_2023}
Shunyu Yao, Dian Yu, Jeffrey Zhao, Izhak Shafran, Thomas~L. Griffiths, Yuan Cao, and Karthik Narasimhan.
\newblock Tree of {Thoughts}: {Deliberate} {Problem} {Solving} with {Large} {Language} {Models}, December 2023.
\newblock arXiv:2305.10601.

\bibitem{zelikman_star_2022}
Eric Zelikman, Yuhuai Wu, Jesse Mu, and Noah~D. Goodman.
\newblock {STaR}: {Bootstrapping} {Reasoning} {With} {Reasoning}, May 2022.
\newblock arXiv:2203.14465.

\bibitem{zhang_instruction_2024}
Shengyu Zhang, Linfeng Dong, Xiaoya Li, Sen Zhang, Xiaofei Sun, Shuhe Wang, Jiwei Li, Runyi Hu, Tianwei Zhang, Fei Wu, and Guoyin Wang.
\newblock Instruction {Tuning} for {Large} {Language} {Models}: {A} {Survey}, November 2024.
\newblock arXiv:2308.10792.

\bibitem{zhou_least--most_2023}
Denny Zhou, Nathanael Schärli, Le~Hou, Jason Wei, Nathan Scales, Xuezhi Wang, Dale Schuurmans, Claire Cui, Olivier Bousquet, Quoc Le, and Ed~Chi.
\newblock Least-to-{Most} {Prompting} {Enables} {Complex} {Reasoning} in {Large} {Language} {Models}, April 2023.
\newblock arXiv:2205.10625.

\bibitem{zhou_instruction-following_2023}
Jeffrey Zhou, Tianjian Lu, Swaroop Mishra, Siddhartha Brahma, Sujoy Basu, Yi~Luan, Denny Zhou, and Le~Hou.
\newblock Instruction-{Following} {Evaluation} for {Large} {Language} {Models}, November 2023.
\newblock arXiv:2311.07911.

\end{thebibliography}
\end{document}